\pdfoutput=1   % arXiv: force pdfLaTeX (PDF figures); keep within the first lines
\documentclass[11pt]{article}

\usepackage[letterpaper,top=1in,bottom=1in,left=1in,right=1in,%
            headheight=14pt,headsep=18pt]{geometry}

\usepackage[T1]{fontenc}
\usepackage{times}

\usepackage{amsmath,amssymb}
\usepackage{graphicx}
\usepackage{adjustbox}
\usepackage{booktabs}
\usepackage{multirow}
\usepackage{xcolor}
\usepackage{microtype}
\usepackage[authoryear,round]{natbib}
\usepackage{hyperref}
\usepackage{url}
\usepackage{xurl}
\usepackage{cleveref}            % load AFTER hyperref
\hypersetup{colorlinks=true,linkcolor=black,citecolor=black,urlcolor=blue,
  pdftitle={Does RoPE Prevent or Degrade Retrieval Heads? A Mechanistic Analysis Across Model Families},
  pdfauthor={Cengizhan Bayram}}

\usepackage{fancyhdr}
\usepackage{titling}
\pretitle{\begin{center}\rule{\textwidth}{1pt}\\[10pt]\LARGE\scshape}
\posttitle{\\[10pt]\rule{\textwidth}{0.4pt}\end{center}\vskip 4pt}
\preauthor{\begin{center}\large}
\postauthor{\par\end{center}}
\predate{\begin{center}\normalsize}
\postdate{\par\end{center}}
\newcommand{\keywords}[1]{\par\vskip 0.3em\noindent{\bfseries Keywords:\ }#1\par\vskip 0.6em}

\newcommand{\LthreeTheta}{500{,}000}
\newcommand{\LthreeHeads}{47}
\newcommand{\LthreeFrac}{4.59\%}
\newcommand{\LthreeD}{0.07}
\newcommand{\LthreeP}{0.9998}
\newcommand{\LthreePartial}{-0.15}

\newcommand{\LtwoTheta}{10{,}000}
\newcommand{\LtwoHeads}{42}
\newcommand{\LtwoFrac}{4.10\%}
\newcommand{\LtwoD}{0.47}
\newcommand{\LtwoP}{0.43}
\newcommand{\LtwoPartial}{0.06}
\newcommand{\LthreeDsd}{0.01}
\newcommand{\LtwoDsd}{0.04}
\newcommand{\QwenDsd}{0.02}
\newcommand{\OlmoDsd}{0.01}

\newcommand{\QwenTheta}{1{,}000{,}000}
\newcommand{\QwenHeads}{59}
\newcommand{\QwenFrac}{7.53\%}
\newcommand{\QwenD}{-0.49}
\newcommand{\QwenP}{0.0003}
\newcommand{\QwenPartial}{-0.21}

\newcommand{\OlmoTheta}{500{,}000}
\newcommand{\OlmoHeads}{87}
\newcommand{\OlmoFrac}{8.50\%}
\newcommand{\OlmoD}{0.50}
\newcommand{\OlmoP}{0.0001}
\newcommand{\OlmoPartial}{0.18}

\newcommand{\KnockRetDrop}{1.00}
\newcommand{\KnockRandDrop}{0.00}
\newcommand{\KnockHeads}{87}        % OLMo-2, contexts {1024,2048,4096}

\newcommand{\PopHeads}{30}
\newcommand{\PopCtx}{4096}
\newcommand{\PopN}{200}
\newcommand{\PopKdims}{16}
\newcommand{\PopBase}{1.00}
\newcommand{\PopLowFreq}{0.885}
\newcommand{\PopHighFreq}{1.00}
\newcommand{\PopLowUtil}{0.985}
\newcommand{\PopHighUtil}{1.00}
\newcommand{\PopRandom}{1.00}

\newcommand{\PopFreqCI}{[-0.16,\,-0.08]}
\newcommand{\PopMcNemarP}{2.4\times10^{-7}}
\newcommand{\PopDiscordant}{23/23}
\newcommand{\DoseKmid}{32}
\newcommand{\DoseLowMid}{0.18}     % low-freq acc at k=32
\newcommand{\DoseRandMid}{0.98}    % random acc at k=32
\newcommand{\DoseKhi}{48}
\newcommand{\DoseLowHi}{0.12}
\newcommand{\KnockQwenHeads}{58}
\newcommand{\KnockQwenMasked}{0.58}
\newcommand{\KnockQwenDrop}{0.42}
\newcommand{\KnockQwenRand}{1.00}

\newcommand{\PopCtrlLowFreq}{1.00}     % low-freq zeroed in layer-matched non-retrieval heads
\newcommand{\PopHeadSpecGap}{0.115}    % retrieval minus control
\newcommand{\PopPplBase}{2.17}
\newcommand{\PopPplLow}{2.19}
\newcommand{\PopPplIncPct}{0.9}        % relative perplexity increase (%)
\newcommand{\PopNiahDropPct}{11.5}     % relative NIAH drop (%)
\newcommand{\PopSpecRatio}{0.08}       % ppl increase / NIAH drop; <0.33 -> retrieval-specific
\newcommand{\PopNseeds}{3}
\newcommand{\PopFreqEffMS}{-0.115}
\newcommand{\PopFreqEffSD}{0.025}
\newcommand{\PopHeadSpecMS}{0.115}
\newcommand{\PopSpecRatioMS}{0.082}
\newcommand{\PopSpecRatioSD}{0.014}
\newcommand{\QwenLowFreq}{0.31}
\newcommand{\QwenFreqEff}{-0.69}
\newcommand{\QwenFreqEffSD}{0.03}
\newcommand{\QwenFreqCtrl}{0.92}

\newcommand{\QwenFreqPplPct}{10}
\newcommand{\QwenCtxLong}{8192}
\newcommand{\QwenLowFreqLong}{0.22}
\newcommand{\QwenFreqEffLong}{-0.78}

\newcommand{\CovOlmo}{36}
\newcommand{\CovMistral}{31}
\newcommand{\QwenLgHeads}{101}
\newcommand{\QwenLgCovLoFE}{-0.045}   % top-30  (30% coverage) -- looks null
\newcommand{\QwenLgCovMidFE}{-0.585}  % top-51  (50%)
\newcommand{\QwenLgCovHiFE}{-0.975}   % top-101 (100%)

\newcommand{\GemmaFreqEff}{-0.45}     % all 45 heads (100%)
\newcommand{\GemmaCtrl}{1.00}
\newcommand{\MisCovLoFE}{-0.005}      % 31% coverage -> null
\newcommand{\MisCovMidFE}{-0.225}     % 62%
\newcommand{\MisCovHiFE}{-0.69}       % 100%
\newcommand{\MfOlmo}{-0.125}
\newcommand{\MfQwen}{-0.72}
\newcommand{\MfQwenLg}{-0.585}
\newcommand{\MfGemma}{-0.195}
\newcommand{\MfMistral}{-0.14}
\newcommand{\QwenArgFE}{-0.925}        % argmax-defined heads
\newcommand{\QwenCopyFE}{-1.0}         % copy-score-defined heads
\newcommand{\QwenCopyOverlap}{0.14}    % top-K Jaccard argmax vs copy
\newcommand{\QwenCopyCtrl}{0.94}
\newcommand{\MisArgFEc}{-0.205}        % argmax (1024 detect, in this run)
\newcommand{\MisArgCtrlc}{0.585}
\newcommand{\MisCopyFE}{-1.0}          % copy-score-defined heads
\newcommand{\MisCopyCtrl}{1.00}        % head-specificity RECOVERS under copy score
\newcommand{\MisCopyOverlap}{0.36}

\newcommand{\AbSeq}{2048}

\newcommand{\AbJaccard}{0.95}
\newcommand{\AbInter}{86}
\newcommand{\AbUnion}{91}
\newcommand{\AbArgmaxRho}{0.90}
\newcommand{\AbMassRho}{0.99}
\newcommand{\AbDeight}{0.54}
\newcommand{\AbDfp}{0.63}

\newcommand{\Bckpts}{45}
\newcommand{\BtokLo}{84}
\newcommand{\BtokHi}{3859}
\newcommand{\BonsetTok}{2014}
\newcommand{\BonsetStep}{480{,}000}
\newcommand{\BbaseHeads}{\raisebox{0.5ex}{\texttildelow}100}
\newcommand{\BpeakHeads}{449}
\newcommand{\ButilHeadR}{-0.75}

\newcommand{\BdipStep}{700{,}000}
\newcommand{\BdipHeads}{112}

\title{Does RoPE Prevent or Degrade Retrieval Heads?\\
       A Mechanistic Analysis Across Model Families}

\author{%
  Cengizhan Bayram\\[2pt]
  \normalsize Independent Researcher\\[2pt]
  \normalsize \texttt{cengizhanbayramtr@gmail.com}}
\date{\today}

\begin{document}
\maketitle

\begin{abstract}
Retrieval heads, attention heads that copy information from earlier context to
the current position, have been proposed as a mechanistic substrate for
long-context recall in transformer language models. Rotary position embeddings
(RoPE) rotate query and key vectors by frequencies that decay with a base
hyperparameter $\theta$, and a natural hypothesis is that this rotation either
\emph{prevents} retrieval heads from forming or \emph{degrades} their function.
We test this hypothesis mechanistically across four open-weight 7--8B models
spanning two attention regimes (multi-head and grouped-query) and a $100\times$
range of RoPE base ($\theta\in[\LtwoTheta,\QwenTheta]$). Using a paired-seed
needle-in-a-haystack protocol that scores \emph{identical} samples across
models, a layer-clustered permutation test that respects the non-independence
of heads, and a causal head-masking knockout, we report four findings.
(i)~Retrieval heads are real and causally necessary: masking the
\KnockHeads{} detected heads collapses NIAH accuracy from $1.00$ to $0.00$
(drop $\KnockRetDrop{}$) while masking an equal number of random heads has
\emph{no} effect (drop $\KnockRandDrop{}$); the dissociation replicates in a
second family (Qwen).
(ii)~Higher $\theta$ is \emph{not} associated with fewer retrieval heads:
in our four-model sample the prevention prediction (fewer heads at higher
$\theta$) does not hold (LLaMA-3.1, $\theta{=}\LthreeTheta$, has \emph{more}
retrieval heads, \LthreeHeads, than LLaMA-2, $\theta{=}\LtwoTheta$, \LtwoHeads),
a directional, confounded refutation of the ``prevention'' hypothesis (H1). (iii)~There is no \emph{universal} ``RoPE degrades retrieval'' law: across four
models the utility--retrieval relationship is inconsistent, Qwen and OLMo show
statistically significant effects in \emph{opposite} directions (Qwen
$d{=}\QwenD$, OLMo $d{=}\OlmoD$; both significant under a layer-clustered test
and Benjamini--Hochberg correction), while the LLaMA family is null. Because OLMo
and LLaMA-3.1 share the \emph{same} $\theta{=}\OlmoTheta$ yet differ, the effect
is not $\theta$-driven. The significant opposite signs are hard to reconcile with
a single universal law, though four models cannot isolate which factor
(architecture, data, or tokenizer) drives the difference, nor establish a
model-family taxonomy.
(iv)~Building on \citet{chiang2025dimension}, who first showed causally that
masking low-frequency dimensions of retrieval heads harms long-context recall, a
controlled population-level patch confirms and sharpens the effect: zeroing the
low-frequency (long-wavelength) RoPE dimensions across retrieval heads degrades
recall \emph{dose-dependently} ($1.00\!\to\!\DoseLowMid$ when $\DoseKmid$ of $128$
dimensions are zeroed, versus $\DoseRandMid$ for the same number of random
dimensions), an effect that is head-specific (no effect in layer-matched
non-retrieval heads) and, at this scale, task-specific. The causal variable is
RoPE's \emph{frequency} axis, not its norm-utility axis. At adequate head coverage
this \emph{direction} holds in all five models we patched (three seeds for
OLMo-2 and Qwen2.5-7B; single-seed for the larger, new-family, and long-context
runs), across four lineages
(OLMo-2, Qwen2.5-7B/14B, Gemma-2, Mistral) and two scales, and strengthens at
longer context; a head-coverage dose-response confirms that fixed-size patches
give false nulls. The conclusion is detector-robust: defining heads by a stricter
teacher-forced copy score instead of our argmax proxy gives the same or a stronger
effect (Qwen) and, for Mistral, restores a clean head-specific control, so its
argmax ``failure'' was a localization artifact. What we do \emph{not} claim is
cross-model \emph{magnitude}, which is confounded by both coverage and detector
localization. We patch five 7--14B models in total. We release all code, a
paired-seed reproducibility harness, and per-checkpoint training-dynamics data,
all available at
\url{https://github.com/CengizhanBayram/Does-RoPE-Prevent-or-Degrade-Retrieval-Heads-A-Mechanistic-Analysis-Across-Model-Families}.
\end{abstract}


% =====================================================================
\section{Introduction}
\label{sec:intro}

Long-context language models are routinely asked to retrieve a specific fact
buried in thousands of tokens, yet \emph{how} they do so is only partly
understood. A leading mechanistic account is the \emph{retrieval head}: a small
set of attention heads that copy a needed token from earlier context to the
current position, and whose ablation collapses long-context recall
\citep{wu2025retrieval}. In parallel, essentially all modern open LLMs encode
position with rotary embeddings (RoPE) \citep{su2024roformer}, which rotate
queries and keys by frequencies set by a base hyperparameter $\theta$; raising
$\theta$ is the standard lever for extending context, and recent work argues that
many RoPE dimensions become low-utility, effectively ``inefficient'', at long
range \citep{chiang2025dimension}.

These two threads invite a question that, to our knowledge, has not been tested
mechanistically: \emph{does RoPE, and its base $\theta$, help or hurt the
retrieval heads long-context recall depends on?} One could argue either way: a
larger $\theta$ might crowd out or destabilise retrieval heads (prevention), or
the low-utility dimensions RoPE induces might be exactly the ones retrieval can
discard without harm (degradation). We test both across four open-weight 7--8B
models spanning two attention regimes and a $100\times$ range of $\theta$, and
find that neither simple story holds: $\theta$ does not prevent retrieval heads,
and the dimensions retrieval actually depends on are the low-\emph{frequency}
ones, not the low-utility ones.

\paragraph{Hypotheses.} We frame two competing hypotheses about how RoPE
interacts with retrieval heads:
\begin{itemize}
  \item \textbf{H1 (Prevention).} Larger $\theta$ (slower-decaying rotation,
  used for long-context models) \emph{reduces the number} of retrieval heads
  that form.
  \item \textbf{H2 (Degradation).} Dimension \emph{utility} (the query-projection
  norm of \citet{chiang2025dimension}) identifies which RoPE dimensions retrieval
  depends on: zeroing the low-utility dimensions leaves recall intact, whereas the
  high-utility ones are load-bearing.
\end{itemize}

\paragraph{Contributions.}
\begin{enumerate}
  \item A \emph{paired-seed} cross-model protocol (\Cref{sec:methods-paired})
  that scores tokenizer-independent samples so model differences cannot be
  attributed to differing inputs.
  \item Results orthogonal to prior work: a $\theta$-versus-head-count test of
  the prevention hypothesis across four models, the training-dynamics emergence
  of retrieval heads in OLMo-2, a whole-head knockout double-dissociation
  replicated in two families, and a quantified, significance-tested account of
  the heterogeneous utility effect (which \citet{chiang2025dimension} noted only
  qualitatively as a Qwen exception).
  \item A controlled, statistically tested replication and extension of
  \citet{chiang2025dimension}'s causal frequency result, adding matched random
  and non-retrieval-head controls, a frequency-aware ordering, a dose-response
  curve, and multi-seed significance, and contrasting the norm and frequency
  framings.
  \item Statistics that avoid pseudoreplication (layer-clustered permutation,
  layer-controlled partial correlation, cross-model FDR) and validate the
  \emph{claim} rather than the detection metric.
  \item An honest, heterogeneous result: retrieval heads are causal, but their
  link to RoPE geometry is family-specific, and H1 is not supported.
\end{enumerate}

% =====================================================================
\section{Related Work}
\label{sec:related}

\paragraph{Retrieval heads.} \citet{wu2025retrieval} identify a small subset of
attention heads that copy a needed token from earlier context to the current
position, and show that masking these heads, but not others, sharply degrades
long-context factuality. They are typically contrasted with the majority of
``streaming'' heads that attend locally \citep{xiao2024duoattention}, and the
retrieval-vs-streaming distinction has become a practical handle on long-context
behaviour, both for memory-efficient inference that keeps only retrieval heads at
full context \citep{xiao2024duoattention} and as a direct optimisation target
\citep{ma2026retmask}. Prior work characterises \emph{that} these heads exist and
matter; our question is the orthogonal one of \emph{how RoPE shapes them}, both
their formation (across $\theta$ and over training) and the dimensions they rely
on. Methodologically we adopt a lighter single-pass
attention-argmax proxy of their copy score and validate it two ways, with a
causal knockout and a teacher-forced copy score (\Cref{sec:methods-detector}).

\paragraph{Mechanistic interpretability of attention heads.} Retrieval heads sit
in a longer line of work that ascribes specific functions to individual heads.
The circuits framework of \citet{elhage2021framework} and the induction heads of
\citet{olsson2022induction} established that heads can implement identifiable
algorithms, induction heads completing $[A][B]\dots[A]\!\to\![B]$ by attending to
the token after a previous occurrence. Retrieval heads are related but distinct:
an induction head keys on local token-level repetition to predict the next token,
whereas a retrieval head copies a semantically required span from far away in the
context to answer a query, and is defined by attention onto a known target rather
than by next-token completion. Our detector and knockout target the latter; we do
not claim our heads are induction heads, and the two need not coincide.

\paragraph{RoPE and its base.} Rotary embeddings \citep{su2024roformer} are
near-universal in open LLMs, LLaMA \citep{touvron2023llama2,grattafiori2024llama3},
Qwen \citep{qwen2024}, OLMo \citep{olmo2024}, in contrast to additive schemes such
as ALiBi \citep{press2022alibi}; increasing the base $\theta$, with the NTK-aware
\citep{bloc97ntk} and YaRN \citep{peng2024yarn} refinements, is the dominant recipe
for context extension. The frequency structure of RoPE has itself drawn scrutiny:
\citet{barbero2025round} analyse which rotary frequencies attention actually uses,
and \citet{du2026rope} prove that at long range RoPE separates neither positions
nor tokens well. This has motivated a family of modifications, partial RoPE that
rotates only some dimensions \citep{khan2026partial}, hybrid RoPE/NoPE attention
\citep{yang2025rope}, dropping positional embeddings post hoc
\citep{gelberg2025drope}, and geometric accounts of long-context RoPE
\citep{wertheimer2026frayed}, all aimed at the same long-range limitations. Our
frequency dissection is complementary: rather than proposing a fix, we causally
locate \emph{which} RoPE dimensions retrieval depends on. Most directly related to
us,
\citet{chiang2025dimension} argue that RoPE drives the dimensions it rotates
through the widest angular range (the high-frequency ones) to low query-projection
utility, and, on the same three models we study (LLaMA-3.1, Qwen-2.5, OLMo-2),
show causally that masking the low-frequency dimensions of the retrieval heads
sharply degrades long-context question answering while masking high-frequency
ones does not. Our Layer-D frequency result (\Cref{sec:layer-d}) is a controlled,
statistically tested replication and extension of theirs: we use a
frequency-aware dimension ordering, add matched random and non-retrieval-head
controls, a dose-response curve, and multi-seed significance, and we test their
norm-utility framing against a frequency framing directly. The remaining
contributions (the $\theta$/head-count test, training dynamics, the whole-head
knockout, and the quantified heterogeneity) are orthogonal to their study.

\paragraph{Architecture and evaluation.} Our models span two attention regimes,
multi-head and grouped-query \citep{ainslie2023gqa}, which we handle explicitly
in the patching hooks. We measure recall with the needle-in-a-haystack protocol
\citep{kamradt2023niah,hsieh2024ruler}, and exploit OLMo-2's released intermediate pretraining
checkpoints \citep{olmo2024} to watch retrieval heads emerge during training.

% =====================================================================
\section{Methods}
\label{sec:methods}

The study has three analyses, which we label by their pipeline stage for brevity:
\emph{Layer~A} (static multi-model detection, \Cref{sec:layer-a}), \emph{Layer~B}
(training dynamics, \Cref{sec:layer-b}), and \emph{Layer~D} (causal validation,
\Cref{sec:layer-d}). The labels are pipeline-stage tags only; there is no separate
Layer~C.

\subsection{Models}
We study four open-weight models chosen to vary the two factors of interest, the
attention regime and the RoPE base, while holding scale roughly fixed at 7--8B:
LLaMA-3.1-8B (GQA, $\theta{=}\LthreeTheta$), LLaMA-2-7B (MHA,
$\theta{=}\LtwoTheta$), Qwen2.5-7B (GQA, $\theta{=}\QwenTheta$), and OLMo-2-7B
(MHA, $\theta{=}\OlmoTheta$). The set (\Cref{tab:models}) spans both attention regimes and a
$100\times$ range of $\theta$, and the LLaMA pair brackets a $50\times$ change of
$\theta$ within one family. OLMo-2 is included specifically because it releases
intermediate pretraining checkpoints (Layer~B), which no other model in the set
provides. All four have $128$-dimensional attention heads. Each model's revision is pinned
to an exact commit hash in our released configuration, and weights are loaded in
8-bit so each model fits a single 24\,GB GPU (an NVIDIA L4 on Google Colab); the
memory-heavy long-context patch ($8192$ tokens, \Cref{sec:layer-d}) and the larger
models (Qwen2.5-14B, Gemma-2-9B) were instead run on a Colab A100 (40\,GB). A
quantization ablation (\Cref{sec:ablation}) confirms the 8-bit results match fp16.
The workload is single-GPU throughout: 8-bit weights keep every 7--14B model
within one card, and each run writes its results to disk before the model is
unloaded, so the pipeline is resumable in short sessions (the heaviest single job
is the Layer-B sweep over 45 OLMo-2 checkpoints; the rest are hours-scale per model).

\begin{table}[t]
\centering
\caption{The four models, chosen to span attention regime and RoPE base at fixed
scale. All have $128$-dimensional heads; each is pinned to a specific commit in
our released config.}
\label{tab:models}
\begin{tabular}{lccccc}
\toprule
Model & Attn. & Layers & Heads & $\theta$ & License \\
\midrule
LLaMA-2-7B   & MHA & 32 & 32 & \LtwoTheta   & Llama-2 \\
LLaMA-3.1-8B & GQA & 32 & 32 & \LthreeTheta & Llama-3.1 \\
Qwen2.5-7B   & GQA & 28 & 28 & \QwenTheta   & Apache-2.0 \\
OLMo-2-7B    & MHA & 32 & 32 & \OlmoTheta   & Apache-2.0 \\
\bottomrule
\end{tabular}
\end{table}

\subsection{Needle-in-a-haystack task}
\label{sec:methods-niah}
Each NIAH sample embeds a short ``needle'' of the form \emph{``The secret
passphrase is \texttt{CODE}.''}, where \texttt{CODE} is five random
alphanumeric characters, at a controlled relative position inside a long
distractor ``haystack'' drawn from PG-19 (public-domain books, with a fixed
neutral fallback corpus if PG-19 is unavailable), followed by the query
\emph{``What is the secret passphrase?''}. Because the needle code is randomly
generated per sample, it cannot appear in any model's pretraining data, so
haystack/training overlap cannot leak the answer; it can at most make the
distractor text more familiar, which would if anything raise the baseline
uniformly across models. We sweep context lengths
$\{1024,2048,4096,8192\}$ and needle positions $\{0.1,0.25,0.5,0.75,0.9\}$;
recall is scored as an exact match of \texttt{CODE} in the generated answer. The
$8192$ length is used only for detection (Layer~A); generation-based experiments
cap at $4096$ because $8192$ exceeds the 8-bit memory budget. Token budgets are
respected so the query is never truncated.

\subsection{Retrieval-head detection}
\label{sec:methods-detector}
We adopt a single-pass attention-argmax proxy of the retrieval-head score of
\citet{wu2025retrieval}. For each needle-in-a-haystack (NIAH) sample we locate
the needle token span and, for every head, measure how often its attention
argmax at the answer position falls on the needle (an \emph{argmax} score); we
also record the total attention \emph{mass} on the needle as a robustness
metric. A head is labelled \emph{retrieval} if its mean score exceeds a
threshold (default $0.1$); \Cref{sec:layer-a} reports robustness across
thresholds. The resulting head \emph{count} is not a fixed quantity: it depends on
the detection context length and the threshold, so it varies modestly across our
runs (for example OLMo $81$--$95$, Qwen $58$--$64$, Mistral $96$--$98$ across
different context sets). We therefore report each experiment's own count
(\Cref{tab:app-provenance} maps every run to its detected counts, so each
main-text number traces to its source), and for the cross-model patches we always
patch a fixed \emph{fraction} of that run's detected set, so the comparison is
coverage-fair regardless of the absolute count.
We emphasise this is an \emph{adapted proxy}, not the original multi-pass
copy-paste metric. We assess its validity carefully, because all
downstream findings rest on the detected head set. (i)~\emph{Functional}: the
selected heads are causally necessary, masking them collapses recall while
masking random heads does not (\Cref{sec:layer-d}); note this establishes
sufficiency, not completeness. (ii)~\emph{Metric}: the per-head proxy scores
correlate only \emph{moderately} with a stricter teacher-forced copy score
(attention from the emitted answer tokens back to the needle) on OLMo
(Spearman $\rho{=}0.54$); the two rank heads similarly but not identically, so
absolute head \emph{identity} is partly proxy-dependent (\Cref{sec:limitations}).
(iii)~\emph{Robustness of the conclusion that matters}: because of (ii), we
verify that the central heterogeneity result does not hinge on the detector. When
heads are re-defined by the copy score instead of the argmax proxy, the utility
effect keeps its sign in both models (OLMo $+0.28\!\to\!+0.45$, Qwen
$-0.58\!\to\!-0.62$). Strikingly, the two detectors share only $22\%$ of Qwen's
retrieval heads (top-$N$ Jaccard; $47\%$ for OLMo), yet \emph{both} still give a
significant negative effect: the sign does not depend on \emph{which} heads are
selected, which strengthens rather than weakens the finding. (Qwen's retrieval is
concentrated in very few heads, $86\%$ of its heads have an exactly-zero
score, so a rank correlation there is degenerate and we rely on the sign test;
for OLMo, whose scores are denser, the rank correlation is $\rho{=}0.54$.)
Absolute head identity is thus proxy-dependent, but the opposite-sign
heterogeneity is \emph{robust to the detector}, the property our claims rely on.

\subsection{RoPE dimension utility and the frequency axis}
\label{sec:methods-utility}
Following \citet{chiang2025dimension}, we proxy the \emph{utility} of each query
dimension by the $L_1$ norm of the corresponding row of the query projection
$W_Q$, the intuition being that a dimension the model has learned to down-weight
contributes little. Separately, every RoPE dimension pair $i$ has an intrinsic
rotation frequency $\theta^{-2i/d_h}$: low-index pairs rotate quickly (high
frequency, short wavelength, sensitive to local offsets) and high-index pairs
rotate slowly (low frequency, long wavelength, the components that remain
distinguishable over long distances). These are two distinct orderings of the
$128$ dimensions, utility (norm magnitude) and frequency (rotation rate), and a
central question of \Cref{sec:layer-d} is which one is causally relevant.
Importantly, the mapping from storage index to frequency is not the identity:
under the \texttt{rotate\_half} (NeoX) convention used by all four models,
dimension $j$ is paired with $j+d_h/2$, so the contiguous ``first/last'' blocks
of raw indices do not coincide with the lowest/highest frequencies. We therefore
compute an explicit frequency ordering and use it (rather than raw index) when we
select the low- and high-frequency dimensions for the causal test.

\subsection{Paired-seed cross-model protocol}
\label{sec:methods-paired}
Different tokenizers segment the same text into different token counts, so a
fixed token-length NIAH sample is not the same task across models. We generate
NIAH \emph{specifications} (haystack text, needle, position) independent of any
tokenizer, then keep only the specifications whose realised token lengths are
valid for \emph{every} model's context budget (an intersection-drop). All
models are therefore scored on an \emph{identical} sample set per seed, so
cross-model differences reflect the model, not the input. We repeat the whole
protocol across seeds $\{42,123,2024\}$ to obtain variance estimates.

\subsection{Statistics}
\label{sec:methods-stats}
Heads within a layer share inputs and are not independent, so a naive per-head
test over-counts evidence (pseudoreplication) and inflates significance. Our
primary test is therefore a \textbf{layer-clustered permutation test}: we permute
the retrieval/non-retrieval labels \emph{within} each layer, preserving the
layer structure, recompute the retrieval-vs-non-retrieval mean utility
difference, and compare the observed value against this null over $10{,}000$
permutations. We complement it with (i)~Cohen's $d$ as a scale-free descriptive
effect size; (ii)~a \textbf{layer-controlled partial Spearman correlation}
between dimension utility and retrieval score, using within-layer demeaning and a
cluster bootstrap to remove the shared layer-depth trend that would otherwise
inflate a raw correlation; (iii)~bootstrap $95\%$ confidence intervals; and
(iv)~Benjamini--Hochberg false-discovery-rate control across the four models
\citep{benjamini1995}. For the paired population-patch comparison
(\Cref{sec:layer-d}) the low- and high-frequency conditions are evaluated on the
\emph{same} samples, so we use an exact \textbf{McNemar} test on their per-sample
correctness together with a bootstrap CI on the paired accuracy difference.
Multi-seed quantities are reported as mean $\pm$ SD over seeds
$\{42,123,2024\}$. The gap between a moderate Cohen's $d$ and a non-significant
clustered $p$ (\Cref{sec:layer-a}) is itself a useful diagnostic that an apparent
effect is pseudoreplicated.

% =====================================================================
\section{Static Multi-Model Analysis (Layer~A)}
\label{sec:layer-a}

\paragraph{Setup.} For each of the four models we run the paired-seed detector
of \Cref{sec:methods-detector} over context lengths
$\{1024,2048,4096,8192\}$ and the standard set of needle positions, label
retrieval heads, compute dimension utility, and test the
retrieval-vs-non-retrieval utility difference with the layer-clustered
permutation test. \Cref{tab:layer-a} summarises the per-model result; effect
sizes and $p$-values are averaged over the three seeds $42/123/2024$, with the
SD of $d$ shown to make the cross-seed stability explicit.

\begin{table}[t]
\centering
\caption{Layer-A retrieval heads and dimension-utility test per model.
$d$ is Cohen's $d$ for the retrieval-vs-non-retrieval utility difference
(mean $\pm$ SD over three seeds, $42/123/2024$); $p$ is the layer-clustered
permutation $p$-value (three-seed mean); head counts, fraction, and
$\rho_{\text{partial}}$ are from the seed-42 paired run. Bold $p$ are significant
at $0.05$ and survive Benjamini--Hochberg correction across the four models.
Qwen has $784$ heads ($28\times28$); the others have $1024$.}
\label{tab:layer-a}
\begin{adjustbox}{max width=\linewidth}
\begin{tabular}{lccccccc}
\toprule
Model & Attn. & $\theta$ & \#Heads & Frac. & $d$ & $p_{\text{clustered}}$ & $\rho_{\text{partial}}$ \\
\midrule
LLaMA-3.1-8B & GQA & \LthreeTheta & \LthreeHeads & \LthreeFrac & $\LthreeD\,{\pm}\,\LthreeDsd$ & \LthreeP & \LthreePartial \\
LLaMA-2-7B   & MHA & \LtwoTheta   & \LtwoHeads   & \LtwoFrac   & $\LtwoD\,{\pm}\,\LtwoDsd$   & \LtwoP   & \LtwoPartial \\
Qwen2.5-7B   & GQA & \QwenTheta   & \QwenHeads   & \QwenFrac   & $\QwenD\,{\pm}\,\QwenDsd$   & \textbf{\QwenP} & \QwenPartial \\
OLMo-2-7B    & MHA & \OlmoTheta   & \OlmoHeads   & \OlmoFrac   & $\OlmoD\,{\pm}\,\OlmoDsd$   & \textbf{\OlmoP} & \OlmoPartial \\
\bottomrule
\end{tabular}
\end{adjustbox}
\end{table}

\paragraph{Finding 1: retrieval heads exist in all families.} Every model forms
a small fraction ($4$--$9\%$) of heads that systematically attend to the
needle (LLaMA-2 \LtwoFrac, LLaMA-3.1 \LthreeFrac, Qwen \QwenFrac, OLMo
\OlmoFrac), replicating the qualitative phenomenon of \citet{wu2025retrieval}
across both MHA and GQA architectures and across a $100\times$ range of $\theta$.

\paragraph{H1 (prevention) is not supported by the observed trend.} If higher
$\theta$ prevented retrieval heads, head count would fall as $\theta$ rises. It
does not (\Cref{tab:layer-a}): the high-$\theta$ models match or exceed the
low-$\theta$ one, LLaMA-3.1 ($\theta{=}\LthreeTheta$, \LthreeHeads{} heads) has
\emph{more} than LLaMA-2 ($\theta{=}\LtwoTheta$, \LtwoHeads), and Qwen
($\theta{=}\QwenTheta$) more still (\QwenHeads). Within the LLaMA family in
isolation, raising $\theta$ $50\times$ \emph{increases} head count
(\LtwoHeads${\to}$\LthreeHeads), the opposite of prevention. We stress that this
is a \emph{directional} observation, not a controlled $\theta$ manipulation: no
two of our models differ in $\theta$ alone (data, tokenizer, and architecture
co-vary), so we cannot attribute the trend to $\theta$ causally. We therefore claim only
that the prevention prediction (fewer heads at higher $\theta$) does not hold in
any of the four models, including within
the LLaMA family (\Cref{sec:limitations}).

\paragraph{Detection is not a grouped-query artifact.} In GQA models several
query heads share key/value projections, which could in principle make a whole
KV group light up together and inflate the retrieval-head count. It does not: the
detected retrieval heads are spread \emph{across} KV groups, not clustered within
them. In Qwen (group size $7$) the active KV groups average $2.1$ retrieval heads
each and only $3.7\%$ are fully retrieval; in LLaMA-3.1 (group size $4$) the
average is $1.5$ and $2.9\%$ are full (\Cref{tab:app-gqa}). So KV sharing does not explain the GQA head
counts, and the cross-architecture comparison is not an artifact of the detector.

\paragraph{Finding 2: the utility--retrieval link is family-specific, and
significant in \emph{opposite} directions.} Qwen retrieval heads have
\emph{lower} dimension utility than non-retrieval heads ($d{=}\QwenD$,
clustered $p{=}\QwenP$), consistent with H2 (low-utility/degradation), whereas
OLMo retrieval heads have \emph{higher} utility ($d{=}\OlmoD$, clustered
$p{=}\OlmoP$), the opposite pattern. Both survive Benjamini--Hochberg correction
across the four models ($2/4$ rejected: Qwen and OLMo; the LLaMA family not).
Across three seeds these effects are remarkably stable, Qwen
$d{=}-0.49\pm0.02$, OLMo $d{=}+0.50\pm0.01$ (mean$\pm$SD), so they are not seed
noise.
The same sign split appears in the layer-controlled partial Spearman correlation
($\rho{=}\QwenPartial$ for Qwen, $\rho{=}\OlmoPartial$ for OLMo). Crucially,
\textbf{OLMo and LLaMA-3.1 share the identical $\theta{=}\OlmoTheta$} yet behave
differently (significant $+0.50$ vs.\ null), so the effect cannot be attributed
to $\theta$. There is no single monotone ``RoPE degrades retrieval'' law. We are
deliberately conservative about what four models can show: the multi-seed CIs
(\Cref{sec:methods-paired}) establish that each model's effect is stable rather
than noise, so significant \emph{opposite-signed} effects demonstrably
\emph{exist} (which refutes universality); but four models cannot establish a
model-family taxonomy, which we leave to a larger model set.

\begin{figure}[t]
\centering
\includegraphics[width=0.62\linewidth]{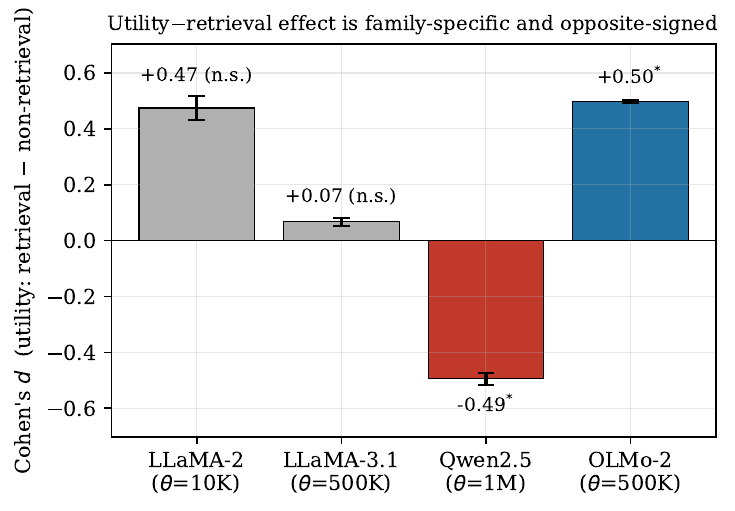}
\caption{Layer-A dimension-utility effect (Cohen's $d$, retrieval vs.\
non-retrieval heads), mean $\pm$ SD over three seeds. Qwen and OLMo are
significant (FDR; ${}^{*}$) and \emph{opposite-signed}; the LLaMA family is not.
OLMo and LLaMA-3.1 share $\theta{=}500$K yet differ, so the effect is not
$\theta$-driven, and the tight SDs show the signs are not seed noise.}
\label{fig:heterogeneity}
\end{figure}

\paragraph{Finding 3: effect size $\neq$ significance (why clustering matters).}
LLaMA-2 illustrates the pseudoreplication trap directly: its Cohen's $d{=}\LtwoD$
is a \emph{moderate} effect with a tiny naive $t$-test $p$ ($7\times10^{-6}$),
yet the layer-clustered permutation test returns $p{=}\LtwoP$ (not significant).
Treating the $1024$ heads as independent would have reported a spurious effect;
respecting within-layer dependence does not. We report the clustered test
throughout, which is precisely what separates the genuine Qwen/OLMo effects from
the spurious LLaMA-2 one.

\paragraph{Robustness.} The findings are stable across detection thresholds:
over $\tau\in[0.05,0.3]$ each model preserves the sign and significance of its
utility effect (Qwen $d\in[-0.56,-0.47]$, all $p<0.002$; OLMo
$d\in[0.22,0.55]$, all $p<0.02$; LLaMA-3.1 null throughout). The direction and
significance of the OLMo effect are further preserved under an 8-bit vs fp16
quantization ablation (\Cref{sec:ablation}). We additionally release a
per-dimension norm diagnostic that \emph{measures} the rotate-half boundary dip
rather than assuming it, so a spurious low-norm dimension cannot be mistaken for
a genuine one.

% =====================================================================
\section{Training Dynamics (Layer~B, OLMo-2)}
\label{sec:layer-b}
OLMo-2 releases intermediate pretraining checkpoints, letting us watch retrieval
heads \emph{form}. We analyse \Bckpts{} stage-1 checkpoints spanning
\BtokLo{}--\BtokHi{}\,B training tokens, recomputing the full Layer-A pipeline at
each.

\begin{figure}[t]
\centering
\includegraphics[width=0.62\linewidth]{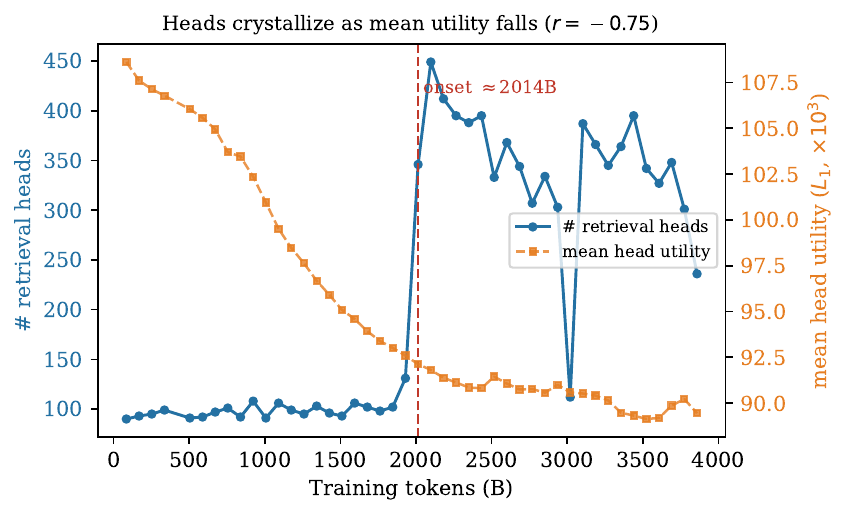}
\caption{Retrieval-head count (left axis) and mean head utility (right axis,
$L_1$ norm) across \Bckpts{} OLMo-2 stage-1 checkpoints. The count is flat
(\BbaseHeads{} heads) until $\sim\BonsetTok$\,B tokens, then rises sharply by
$\sim3.5\times$ to a $300$--\BpeakHeads{} plateau; mean utility falls as the count
rises (Pearson $r{=}\ButilHeadR$). The transient dip near step~\BdipStep{} is
correctly ignored by the midpoint-crossing onset detector. Single OLMo-2 run.}
\label{fig:layerb}
\end{figure}

\paragraph{Retrieval heads crystallize abruptly.} The retrieval-head count is
flat and low (\BbaseHeads{} heads) for the first \BonsetTok{}\,B tokens, then
rises sharply by roughly $3.5\times$ to a plateau of $300$--\BpeakHeads{} heads.
A midpoint-crossing onset detector (robust to transient spikes) places the
crystallization onset at step~\BonsetStep{} ($\BonsetTok$\,B tokens). We describe
this as \emph{phase-transition-like} purely descriptively (an abrupt onset rather
than a gradual drift, \Cref{fig:layerb}); we do not claim a formal phase
transition, and because these are checkpoints from a \emph{single} OLMo-2
pretraining run, we report the abruptness as an observation that may depend on the
optimizer, data mixture, and learning-rate schedule of that run
(\Cref{sec:limitations}). Over the same checkpoints, mean head utility falls as
the count rises (Pearson $r{=}\ButilHeadR$, \Cref{fig:layerb}); we report this
co-movement descriptively and make no claim about temporal ordering, since the
two series are autocorrelated.

\paragraph{Dimension utility tracks head formation.} Across checkpoints, mean
query-projection utility is strongly anti-correlated with the retrieval-head
count (Pearson $r{=}\ButilHeadR$): as retrieval heads proliferate, the mean
query-projection norm falls. We deliberately make \emph{no} claim about temporal
ordering (which leads which). The series are short ($\Bckpts$ checkpoints) and
strongly autocorrelated, so a lead--lag permutation test would be
anti-conservative, and a single training run cannot separate ``utility leads
heads'' from ``heads lead utility.'' We report the association, not its
direction.

\paragraph{Robustness.} A sharp transient dip at step~\BdipStep{}
(\BdipHeads{} heads, between plateau values of $300$--$400$) is correctly
\emph{ignored} by the midpoint-crossing onset detector; a naive
$\arg\max$-of-difference detector would have misfired on this recovery spike.

% =====================================================================
\section{Causal Validation and the RoPE-Frequency Axis (Layer~D)}
\label{sec:layer-d}

\paragraph{Head-masking knockout (completed).} To confirm that the detected
heads are causally \emph{necessary} for recall, not merely correlated with
it, we ablate them and measure NIAH accuracy. A head is ablated by zeroing its
output through a forward hook (model weights are never modified), and recall is the
exact-match rate of the five-character passphrase over the $50$ samples per context.
Because scoring is an exact string match, a model that has lost the copy circuit
cannot emit the passphrase (score $0$) while the matched random control preserves it
(score $1$), so the dissociation is near-binary by construction, not a tuned outcome.
On OLMo-2 (the model with the
most retrieval heads, \KnockHeads{}), over contexts $\{1024,2048,4096\}$,
masking the retrieval heads collapses accuracy from a perfect baseline to zero
(mean accuracy $1.00\!\rightarrow\!0.00$, drop $\KnockRetDrop{}$), while masking
an equal number ($\KnockHeads{}$) of randomly chosen non-retrieval heads leaves
accuracy untouched (drop $\KnockRandDrop{}$; \Cref{tab:knockout}). This double dissociation, total
collapse for retrieval heads, zero effect for the matched random control, is
the strongest causal evidence in the paper and rules out a coincidental
correlation between the detection score and recall. The dissociation replicates
in a second family: on Qwen2.5 (GQA, $\KnockQwenHeads{}$ retrieval heads) masking
the retrieval heads drops recall from $1.00$ to $\KnockQwenMasked{}$
(drop~$\KnockQwenDrop{}$) while masking the same number of random heads leaves it
at $\KnockQwenRand{}$ (drop~$0.00$), a clear, if partial, collapse versus the
total collapse in OLMo. This whole-head ablation removes the mechanism outright and
should be read separately from the graded dimension-level patch of \Cref{tab:dose}
(zeroing only 16 of 128 RoPE dimensions, which lowers recall to $0.885$): the two
differ in kind, deletion versus partial degradation, not merely in degree.

\begin{table}[t]
\centering
\caption{Head-masking knockout (NIAH accuracy). Masking the detected retrieval
heads collapses recall; masking an equal number of random non-retrieval heads
does not. The double dissociation holds in two attention families (total in
OLMo, partial in Qwen). Detected head counts are context- and threshold-dependent,
so they differ across tables; each table reports its own run's count
(\Cref{sec:methods-detector}).}
\label{tab:knockout}
\begin{tabular}{lcccc}
\toprule
Model & \#Ret.\ heads & Baseline & Retrieval-masked & Random-masked \\
\midrule
OLMo-2-7B (MHA)  & \KnockHeads      & 1.00 & 0.00 ($-1.00$) & 1.00 ($-0.00$) \\
Qwen2.5-7B (GQA) & \KnockQwenHeads  & 1.00 & \KnockQwenMasked\ ($-\KnockQwenDrop$) & 1.00 ($-0.00$) \\
\bottomrule
\end{tabular}
\end{table}

\paragraph{Per-head zeroing is uninformative at ceiling.} We first tried the
single-head test of \citet{chiang2025dimension}'s logic: for each retrieval head,
zero its $\PopKdims$ lowest/highest-utility (and lowest/highest-frequency)
dimensions and measure recall. At the longest context that fits in 8-bit on a
24\,GB GPU ($\PopCtx$ tokens) OLMo solves NIAH at ceiling ($\text{accuracy}=1.00$),
and zeroing $\PopKdims$ dimensions of a \emph{single} head never moves it: all
conditions return $1.00$ for all \PopHeads{} top heads. This is not ``no causal
effect'' but ``insufficient leverage at ceiling'': one head out of many has too
little influence to overcome the model's margin. A meaningful causal test must
either escape the ceiling or apply more leverage. For the population test we
therefore patch the \PopHeads{} heads with the highest argmax retrieval score (a
high-precision subset of the $\sim$84 detected in OLMo): patching the strongest
heads maximises the intervention's leverage while bounding cost, and the
non-retrieval control is drawn from the same layers so the comparison is matched.

\paragraph{Population-level frequency patching (the \S6 test).} This test
revisits, with added controls and statistics, the causal masking of
\citet{chiang2025dimension}, who found that masking the low-frequency dimensions
of retrieval heads degrades long-context QA. We therefore patch the \emph{same
dimension class across all \PopHeads{} top retrieval heads simultaneously} and
compare conditions on a shared set of $\PopN$ NIAH samples at $\PopCtx$ tokens.
The result is sharply frequency-specific
(\Cref{tab:layer-d-pop}). Zeroing the $\PopKdims$ \emph{lowest-frequency}
(long-wavelength) RoPE dimensions across all heads drops recall to
$\PopLowFreq$, whereas zeroing the $\PopKdims$ highest-frequency dimensions, an
equal number of random dimensions, or the highest-utility ($L_1$-norm)
dimensions leaves recall at ceiling ($\geq\PopLowUtil$). The low-frequency vs
high-frequency contrast is paired (same samples) and significant: an exact
McNemar test gives $p=\PopMcNemarP$ with all discordant pairs one-sided
($\PopDiscordant$), and the bootstrap CI on the accuracy difference,
$\PopFreqCI$, excludes zero (\Cref{tab:layer-d-pop} reports this representative
run). The effect replicates across all $\PopNseeds$ seeds ($42/123/2024$): the
frequency effect is $\PopFreqEffMS\pm\PopFreqEffSD$ at $k{=}\PopKdims$, negative
and McNemar-significant in every seed ($p\le 8\times10^{-6}$). This $k{=}\PopKdims$
figure is the conservative end of the dose-response below.

\begin{table}[t]
\centering
\caption{Population-level patching on OLMo-2 (top-\PopHeads{} retrieval heads,
which is $\sim\!\CovOlmo\%$ of OLMo's $\sim\!84$ detected heads, $\PopCtx$-token
context, $\PopN$ samples, $\PopKdims$ dims zeroed per head, all heads patched
simultaneously; head coverage is swept in \Cref{tab:cov-sweep}). Only zeroing the
lowest-\emph{frequency} dimensions degrades recall; the utility ($L_1$-norm) axis
is null. The $\sim\!84$ count is this run's; detection is context/threshold-dependent,
so counts differ across tables (\Cref{sec:methods-detector}).}
\label{tab:layer-d-pop}
\begin{tabular}{lc}
\toprule
Condition (dims zeroed across all heads) & NIAH accuracy \\
\midrule
baseline (no patch)        & \PopBase \\
highest-utility ($L_1$ norm) & \PopHighUtil \\
random                     & \PopRandom \\
lowest-utility ($L_1$ norm)  & \PopLowUtil \\
highest-frequency (RoPE)   & \PopHighFreq \\
\textbf{lowest-frequency (RoPE)} & \textbf{\PopLowFreq} \\
\midrule
lowest-frequency, in layer-matched \emph{non-retrieval} heads (control) & \PopCtrlLowFreq \\
\bottomrule
\end{tabular}
\end{table}

\paragraph{Dose-response.} The $k{=}\PopKdims$ result above is the conservative
end of a monotone dose-response (\Cref{tab:dose}). As more low-frequency
dimensions are zeroed across the retrieval heads, recall falls steeply, to
$\DoseLowMid$ at $k{=}\DoseKmid$ and $\DoseLowHi$ at $k{=}\DoseKhi$ (of $128$
per-head dimensions), while zeroing the \emph{same number} of random dimensions
barely moves it ($\DoseRandMid$ at $k{=}\DoseKmid$). The effect is therefore not
marginal: once enough low-frequency dimensions are removed the model essentially
cannot retrieve, and the gap to the random control widens with $k$. This
dose-dependence, with a matched random control at every step, is strong evidence
that retrieval genuinely depends on the low-frequency RoPE dimensions.

\begin{table}[t]
\centering
\caption{Dose-response on OLMo-2 (top-\PopHeads{} heads, $\PopCtx$ tokens):
NIAH accuracy as $k$ lowest-frequency vs $k$ random dimensions are zeroed across
all heads. Low-frequency removal collapses recall dose-dependently; matched
random removal does not.}
\label{tab:dose}
\begin{tabular}{rccc}
\toprule
$k$ & low-freq acc. & random acc. & low-freq drop \\
\midrule
8  & 0.985 & 1.00 & 0.015 \\
16 & 0.885 & 1.00 & 0.115 \\
32 & \textbf{0.175} & 0.98 & \textbf{0.825} \\
48 & 0.115 & 0.87 & 0.885 \\
64 & 0.070 & 0.375 & 0.930 \\
\bottomrule
\end{tabular}
\end{table}

\begin{figure}[t]
\centering
\includegraphics[width=0.62\linewidth]{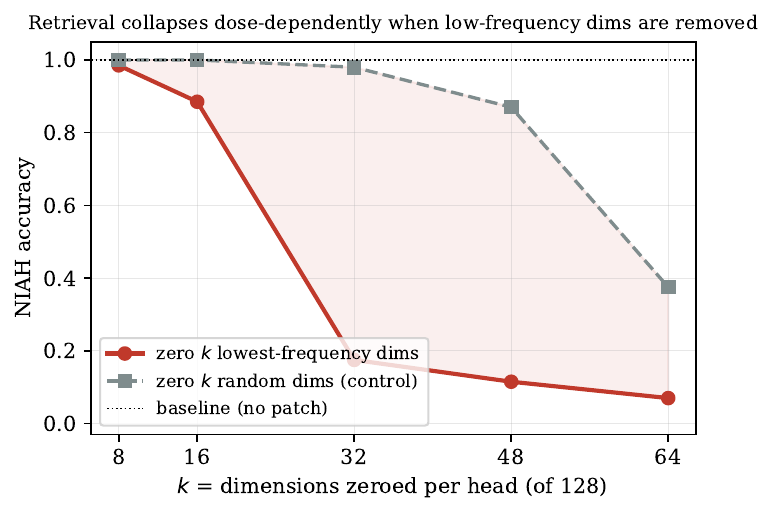}
\caption{Dose-response on OLMo-2 (top-\PopHeads{} heads, $\PopCtx$ tokens).
Zeroing the $k$ lowest-frequency RoPE dimensions across the retrieval heads
collapses NIAH recall as $k$ grows (red), while zeroing the \emph{same number} of
random dimensions barely moves it (grey). At $k{=}\DoseKmid$ recall is
$\DoseLowMid$ vs.\ $\DoseRandMid$ for the random control.}
\label{fig:dose}
\end{figure}

\paragraph{Specificity controls.} Two controls, specified in advance (the rule
below was fixed before we saw the perplexity numbers; we make no formal
pre-registration claim), confirm the effect is specific to retrieval rather than a
generic consequence of removing low-frequency dimensions. \emph{(i)~Head-specificity.} Zeroing the same
low-frequency dimensions in an equal number of \emph{layer-matched non-retrieval}
heads leaves recall at ceiling ($\PopCtrlLowFreq$); the retrieval-vs-control gap
is $\PopHeadSpecGap$. The effect therefore requires the dimensions to sit in
retrieval heads, and is not explained by layer depth (controls are drawn from
the same layers). \emph{(ii)~Task-specificity.} Under the identical low-frequency
patch, perplexity on plain $\PopCtx$-token text (no needle) rises only from
$\PopPplBase$ to $\PopPplLow$ ($+\PopPplIncPct\%$), against an $\PopNiahDropPct\%$
relative drop in NIAH recall, a ratio of $\PopSpecRatio$. We label the effect
retrieval-specific by a heuristic cutoff of $0.33$ on this ratio, fixed before we
saw the perplexity numbers; this cutoff is our own convention, not a value from
prior literature, so we report the raw quantities ($+\PopPplIncPct\%$ perplexity
rise vs $\PopNiahDropPct\%$ recall drop) and readers may apply their own threshold.
Both controls hold across all $\PopNseeds$ seeds: the non-retrieval control stays
at ceiling ($\PopCtrlLowFreq$ in every seed; head-specificity gap
$\PopHeadSpecMS\pm\PopFreqEffSD$), and the perplexity ratio is essentially flat
($1.009\pm0.0004$), giving a specificity ratio of $\PopSpecRatioMS\pm\PopSpecRatioSD$
($<0.33$ in every seed). The low-frequency dimensions are thus load-bearing for
long-range \emph{retrieval} specifically, not for general long-context language
modelling.

\paragraph{Replication in Qwen2.5 (a second, grouped-query family).} Running the
identical controlled population patch on Qwen2.5 (GQA), across the same three
seeds, reproduces the core effect and makes it larger. Qwen also solves the
$\PopCtx$-token task at ceiling (baseline $1.00$, so the test is interpretable),
and zeroing the $\PopKdims$ lowest-frequency dimensions of its top-\PopHeads{}
retrieval heads collapses recall to $\QwenLowFreq$ (frequency effect
$\QwenFreqEff\pm\QwenFreqEffSD$, the \emph{three-seed mean}; McNemar $p<10^{-39}$
in every seed; per-seed in \Cref{tab:app-layerd-qwen}), versus near-ceiling for the
high-frequency, random, and highest-norm conditions. The \emph{seed-42} run alone
is $\MfQwen$, and that single-seed value is the one we reuse at matched coverage
(\Cref{tab:xfam-freq}) and as the $4096$ reference in the long-context comparison
below; throughout we label these two as $\QwenFreqEff$ (three-seed mean) and
$\MfQwen$ (seed-42). So the low-frequency dependence is \emph{not} OLMo-specific; it holds
in a grouped-query model too, the headline ``frequency, not norm'' result of
this section.

The magnitude and specificity, however, differ from OLMo in ways we report
plainly. (i)~\emph{We do not attribute the larger effect to GQA.} The frequency
effect is about six times OLMo's at the same $k$ ($\QwenFreqEff$ vs
$\PopFreqEffMS$, both three-seed means), but we cannot isolate the cause: Qwen has a larger RoPE base
($\theta{=}\QwenTheta$) \emph{and} a more concentrated retrieval distribution
(most of its heads score zero, \Cref{sec:methods-detector}), and only one of our
two GQA models was frequency-patched, so the architecture cannot be separated
from these. Notably Qwen is not uniformly more fragile, its head-masking knockout
was \emph{milder} than OLMo's ($1.00\!\to\!\KnockQwenMasked$ vs
$1.00\!\to\!0.00$), which already argues against a simple ``Qwen is just easier to
break'' reading. A further uncontrolled factor is patch \emph{coverage}: the
top-\PopHeads{} heads are a larger fraction of Qwen's retrieval heads than of
OLMo's. We cannot separate these candidates here and return to the magnitude
question, with a third family, below. (ii)~\emph{Task-specificity is only partial in Qwen.} The
low-frequency patch raises plain-text perplexity by $+\QwenFreqPplPct\%$, against
$+\PopPplIncPct\%$ in OLMo, an order of magnitude more: in Qwen these dimensions
are load-bearing for general language modelling, not retrieval alone (directly
echoing the GSM8k finding of \citet{chiang2025dimension}). The specificity ratio
stays below $0.33$ only because the NIAH drop is so large; we therefore call the
effect retrieval-\emph{dominant} in Qwen, not retrieval-specific as in OLMo.
(iii)~\emph{Head-specificity is strong but not absolute.} Zeroing the same
dimensions in layer-matched non-retrieval heads leaves recall at $\QwenFreqCtrl$,
a small ($0.08$) drop, versus an exact $1.00$ in OLMo; the retrieval-vs-control
gap is still large ($0.61$), but Qwen shows a slight non-retrieval effect that
OLMo did not. Taken together, the replication is strong (the frequency axis is
causal in both families) while the specificity is clean in OLMo and partial in
Qwen.

\paragraph{Long context, more families, and a coverage dose-response.} Several
further runs sharpen the scope and, importantly, resolve the magnitude question
(\Cref{tab:xfam-freq}, \Cref{tab:cov-sweep}; full per-run detail in
\Cref{tab:app-extra}).

\emph{Long context.} Pushing the Qwen patch to $\QwenCtxLong$ tokens (single seed,
the \emph{same} top-\PopHeads{} heads, so coverage is fixed) makes the effect
\emph{larger}: recall falls to $\QwenLowFreqLong$ (frequency effect
$\QwenFreqEffLong$ at $8192$ vs the seed-42 value $\MfQwen$ at $4096$; both seed-42,
same top-\PopHeads{}), with baseline still $1.00$ and the control back at $1.00$.
Coverage and seed being identical across contexts, this within-model comparison is
not confounded: the dependence genuinely grows with context.

\emph{Coverage is a second dose-response, and it explains the apparent nulls.}
We had flagged that a fixed top-\PopHeads{} patch covers a different fraction of
each model's retrieval heads. Coverage sweeps confirm this directly and turn it
into a finding (\Cref{tab:cov-sweep}). On Qwen2.5-14B ($\QwenLgHeads$ heads) the
effect grows from a near-null $\QwenLgCovLoFE$ at $30\%$ coverage to
$\QwenLgCovMidFE$ at $50\%$ and a near-total $\QwenLgCovHiFE$ at $100\%$; on
Mistral-7B it grows from $\MisCovLoFE$ (null) at $\CovMistral\%$ to $\MisCovMidFE$
at $62\%$ and $\MisCovHiFE$ at $100\%$. The effect scales monotonically with how
much of the retrieval-head population is patched, a head-coverage dose-response
parallel to the dimension-$k$ one, and a positive control for false nulls: at
$\sim\!30\%$ coverage even models with a large full-coverage effect look null.
Mistral's earlier null was thus \emph{under-coverage}, now directly confirmed: its
low-frequency dependence is real and large once enough of its heads are patched.

\emph{Coverage-matched, the direction is universal.} Patching the same $50\%$ of
every model's argmax-detected heads (\Cref{tab:xfam-freq}) gives a significant
negative frequency effect in all five models across four lineages: OLMo-2
($\MfOlmo$), Qwen2.5-7B ($\MfQwen$) and -14B ($\MfQwenLg$), Gemma-2-9B
($\MfGemma$, a new family), and Mistral-7B ($\MfMistral$). The \emph{direction} is
therefore universal. Two of these five entries are not independent runs: the
Qwen2.5-14B value reuses the $50\%$ point of the coverage sweep
(\Cref{tab:cov-sweep}) and the Qwen2.5-7B value reuses the seed-42 run of the
three-seed patch (\Cref{tab:app-layerd-qwen}); they are listed here only to read
all five models at one matched coverage, not as separate evidence. The apparent
magnitude spread, and the one apparent
specificity failure (Mistral's leaky control), turn out to depend heavily on
detection quality, as we show next, so we do not read them as model properties.

\emph{The conclusion is detector-robust, and a stricter detector cleans it up.}
Our main runs use the single-pass argmax detector, which overlaps only partially
with a teacher-forced copy score (\Cref{sec:methods-detector}). We therefore
re-ran the $50\%$ frequency patch with heads defined by the \emph{copy} score
instead, on the two models where it matters most (single seed; \Cref{tab:copyhead}).
The effect is robust to, in fact strengthened by, the change. In Qwen the
copy-defined heads share only $\QwenCopyOverlap$ of the argmax set, yet give a
\emph{total} collapse ($\QwenCopyFE$ vs $\QwenArgFE$ under argmax): the result is
not an artifact of the argmax proxy. In Mistral the copy detector \emph{resolves}
the earlier specificity failure: the same patch collapses recall completely
($\MisCopyFE$) with a \emph{perfect} non-retrieval control ($\MisCopyCtrl$),
versus a weak, leaky $\MisArgFEc$ at control $\MisArgCtrlc$ under argmax. Mistral's
apparent diffuseness was thus an argmax \emph{localization} failure, not a real
property: with heads identified properly, Mistral too shows a clean, head-specific
frequency dependence. By the same token the cross-model magnitude gap shrinks
under the better detector (Qwen and Mistral both reach total collapse at $50\%$
coverage), so we treat magnitude as confounded by detector localization as well as
by coverage, and claim only the direction across models. A copy-score sweep across
all five models is the clean way to settle magnitude, and we leave it to future
work. The broader lesson is the one we pre-committed to: validate the
\emph{claim}, not the metric, the frequency-specific, head-specific conclusion
holds under both detectors, and the stricter one only sharpens it.

\emph{A new family shows the clean effect.} Gemma-2-9B is a third lineage with
$256$-dim heads, so we scale the dose to a constant fraction of head width,
$k{=}d_h/8$ ($32$ of $256$ dims, the same $12.5\%$ as $16$ of $128$ elsewhere);
note this is the conservative choice, a fixed $k{=}16$ would zero only $6\%$ of
Gemma's head and so \emph{understate} the effect rather than inflate it. With this
dose Gemma shows a clear and \emph{head-specific} frequency effect ($\GemmaFreqEff$
at full coverage, McNemar $p<10^{-12}$, control $\GemmaCtrl$), so the
retrieval-head-specific result is not OLMo/Qwen-only.

\emph{Summary.} At adequate coverage the low-frequency \emph{direction} holds in
every model we tested (five models, four lineages, two scales); apparent nulls are
coverage artifacts (the head-coverage sweeps reproduce them on demand), and the
one apparent specificity failure is a detector artifact (Mistral is clean and
head-specific once heads are defined by the copy score). The frequency-specific,
head-specific conclusion is thus robust to both coverage and detector choice. What
we do \emph{not} claim is cross-model \emph{magnitude}: it is confounded by both
coverage and detector localization (under the copy detector Qwen and Mistral both
collapse fully), so the universal claim is the direction of the effect, not its
size.

\begin{table}[t]
\centering
\caption{Head-coverage dose-response: frequency effect vs.\ the fraction of
detected retrieval heads patched (single seed; dose $k{=}d_h/8$). In both models
the effect is near-null at $\sim\!30\%$ coverage and grows to near-total at full
coverage, so a fixed small patch under-counts it and can read as a false null.}
\label{tab:cov-sweep}
\begin{tabular}{lcccc}
\toprule
Model & \#ret. & coverage & top-$K$ & freq.\ eff. \\
\midrule
Qwen2.5-14B & 101 & 30\%  & 30  & $\QwenLgCovLoFE$ \\
            &     & 50\%  & 51  & $\QwenLgCovMidFE$ \\
            &     & 100\% & 101 & $\QwenLgCovHiFE$ \\
\addlinespace
Mistral-7B  & 97  & 31\%  & 30  & $\MisCovLoFE$ \\
            &     & 62\%  & 60  & $\MisCovMidFE$ \\
            &     & 100\% & 97  & $\MisCovHiFE$ \\
\bottomrule
\end{tabular}
\end{table}

\begin{table}[t]
\centering
\caption{Frequency patch at \emph{matched} $50\%$ head coverage, on
\emph{argmax}-detected heads (single seed; dose $d_h/8$; baseline $1.00$ each).
The direction (negative, significant) holds in all five models over four lineages.
``ctrl'' is recall when the same dimensions are zeroed in matched non-retrieval
heads. The magnitude spread and Mistral's leaky control (ctrl $=0.00$: the non-retrieval patch \emph{also}
collapses recall, so a clean head-specific control would instead sit near $1.00$) are largely
artifacts of argmax localization: under a teacher-forced copy-score detector they
close up (Mistral control returns to $1.00$ and its effect to total collapse,
\Cref{tab:copyhead}). ``\#ret'' is this run's detected count, which is
context/threshold-dependent and so differs from other tables
(\Cref{sec:methods-detector}). The Qwen2.5-14B row is the $50\%$ point of the
coverage sweep (\Cref{tab:cov-sweep}) and the Qwen2.5-7B row is the seed-42 run of
the three-seed patch (\Cref{tab:app-layerd-qwen}); these are the same measurements
read at matched coverage, not independent runs.
$^{\dagger}$Mistral's ctrl${=}0.00$ is an argmax-localization artefact, not a real
leak: under the copy-score detector it returns to $1.00$ (\Cref{tab:copyhead}).}
\label{tab:xfam-freq}
\begin{tabular}{llcccc}
\toprule
Model & Attn. & \#ret. & freq.\ eff. & ctrl & McNemar $p$ \\
\midrule
OLMo-2-7B   & MHA & 95  & $\MfOlmo$    & 1.00 & $6\times10^{-8}$ \\
Qwen2.5-7B  & GQA & 62  & $\MfQwen$    & 0.89 & $9\times10^{-44}$ \\
Qwen2.5-14B & GQA & 101 & $\MfQwenLg$  & 1.00 & $1\times10^{-35}$ \\
Gemma-2-9B  & GQA & 45  & $\MfGemma$   & 1.00 & $4\times10^{-12}$ \\
Mistral-7B  & GQA & 97  & $\MfMistral$ & $0.00^{\dagger}$ & $7\times10^{-9}$ \\
\bottomrule
\end{tabular}
\end{table}

\begin{table}[t]
\centering
\caption{Frequency patch with heads defined by the \emph{argmax} proxy vs.\ the
teacher-forced \emph{copy} score (matched $50\%$ coverage, single seed,
patch at $4096$). ``ovl'' is the top-$K$ Jaccard between the two head sets. Despite
small overlap, the copy detector gives the \emph{same or stronger} effect, so the
frequency result is not a proxy artifact; and for Mistral the copy detector
restores a clean head-specific control ($1.00$), showing its argmax ``failure''
was a localization, not a model, property.}
\label{tab:copyhead}
\begin{tabular}{lccc}
\toprule
Model & ovl & freq.\ eff.\ (argmax / copy) & ctrl (argmax / copy) \\
\midrule
Qwen2.5-7B & $\QwenCopyOverlap$ & $\QwenArgFE$ / $\QwenCopyFE$ & $0.99$ / $\QwenCopyCtrl$ \\
Mistral-7B & $\MisCopyOverlap$  & $\MisArgFEc$ / $\MisCopyFE$  & $\MisArgCtrlc$ / $\MisCopyCtrl$ \\
\bottomrule
\end{tabular}
\end{table}

\paragraph{Interpretation.} Two axes dissociate. The \emph{utility} axis is
causally null here: zeroing the highest-$L_1$-norm dimensions does nothing, so
norm-utility does not identify load-bearing dimensions for retrieval. The
\emph{frequency} axis is causal across the models we tested and, by the controls
above, retrieval-head-specific in OLMo, Qwen, and Gemma, and in Mistral too once
its heads are identified by the copy score rather than the argmax proxy: the
low-frequency (long-wavelength) dimensions
that encode long-range position are the ones retrieval depends on, consistent with
the view that RoPE's slow-rotating dimensions carry the long-distance signal a
needle-in-a-haystack lookup requires.
Thus retrieval's dependence on RoPE geometry runs through the \emph{frequency}
axis, not the \emph{norm-utility} axis, a refinement of the dimension-inefficiency
account toward the frequency axis.

\paragraph{Scope and caveats.} Two caveats bound this result. First, the
perplexity shift is marginal but \emph{nonzero} ($+\PopPplIncPct\%$); the patch
is not perfectly inert on general text, only far below the NIAH drop. Second,
the clean head-\emph{specific} version is established in four families (OLMo-2 and
Qwen2.5 across $\PopNseeds$ seeds, Gemma-2 and Mistral-7B single-seed, the latter
once heads are defined by the copy score, \Cref{tab:copyhead}); cross-model
magnitude is confounded by coverage and detector localization, so the cross-model
claim is the direction (\Cref{sec:layer-d}). Magnitude itself is not a caveat for
the direction:
the dose-response (\Cref{tab:dose}) shows the effect is small only at small $k$
and becomes near-total by $k{=}\DoseKhi$. The graded dimension-zeroing should
still be distinguished from the all-or-nothing head-masking knockout: removing
whole heads deletes the mechanism, whereas zeroing dimensions degrades it
dose-dependently.

% =====================================================================
\section{Quantization Ablation}
\label{sec:ablation}
All models run in 8-bit to fit a 24\,GB GPU. Because the detector is
argmax-based (discrete), a small continuous shift from 8-bit rounding could in
principle flip a head whose top-two positions are close. We therefore re-run
detection in fp16 on OLMo-2 (the model with the most retrieval heads and a
significant utility effect) at seq=$\AbSeq$, and compare at three levels
(\Cref{tab:ablation}). Head identity is nearly unchanged (Jaccard
$\AbJaccard$, $\AbInter/\AbUnion$ heads shared); per-head scores are almost
perfectly rank-correlated (argmax $\rho{=}\AbArgmaxRho$, attention-mass
$\rho{=}\AbMassRho$); and the headline finding, OLMo retrieval heads having
\emph{higher} utility, keeps its sign and significance in both precisions
(Cohen's $d{=}\AbDeight$ in 8-bit vs $d{=}\AbDfp$ in fp16, clustered
permutation $p<10^{-4}$ for both). The finding is therefore not a quantization
artifact.

\begin{table}[t]
\centering
\caption{Quantization ablation on OLMo-2 (8-bit vs fp16, seq=$\AbSeq$). The
\emph{finding} (direction + significance), not byte-identical head sets, is what
is defended.}
\label{tab:ablation}
\begin{tabular}{lll}
\toprule
Level & Metric & Result \\
\midrule
Head identity   & Jaccard of retrieval-head sets & $\AbJaccard$ ($\AbInter/\AbUnion$) \\
Score agreement & Spearman $\rho$ (argmax / mass) & $\AbArgmaxRho$ / $\AbMassRho$ \\
Finding         & Cohen's $d$ (8-bit / fp16)     & $\AbDeight$ / $\AbDfp$, both $p<10^{-4}$ \\
\bottomrule
\end{tabular}
\end{table}

% =====================================================================
\section{Discussion}
\label{sec:discussion}

\paragraph{What the experiments answer.} We posed two intuitive accounts of how
RoPE shapes retrieval and found both wrong in their stated form. Prevention (H1)
predicts that the slower-decaying rotation of a larger base $\theta$ should
suppress retrieval-head formation; instead the number of retrieval heads
\emph{rises} with $\theta$ across four models (\Cref{tab:layer-a},
\Cref{fig:heterogeneity}), and even within the LLaMA family the high-$\theta$
model carries more heads than the low-$\theta$ one. Utility-degradation (H2)
predicts that the dimensions flagged as low-utility by query-projection norm are
causally inert while high-utility ones carry retrieval; instead the norm-utility
axis is causally \emph{null}, since zeroing the highest-norm dimensions leaves
recall at ceiling (\Cref{tab:layer-d-pop}). What remains, and what we argue is
the correct picture, is that retrieval depends on RoPE's \emph{frequency} axis:
the low-frequency, long-wavelength dimensions that encode long-range position.

\paragraph{Why the frequency axis is load-bearing.} RoPE assigns each dimension
pair a rotation frequency $\theta^{-2i/d_h}$, so low-index pairs rotate quickly
(short wavelength, sensitive to local offsets) while high-index pairs rotate
slowly (long wavelength, the components that stay distinguishable across
thousands of tokens). A needle-in-a-haystack lookup is exactly the operation
that must relate a query position to a key thousands of tokens away, so it can
only succeed by reading the dimensions whose phase has not wrapped around over
that distance, that is, the low-frequency ones. Our causal result is the
mechanistic confirmation: zeroing the low-frequency dimensions across the
retrieval heads collapses recall (\Cref{fig:dose}), whereas zeroing an equal
number of high-frequency or random dimensions does not. Retrieval is sensitive
not to \emph{how much} a dimension is used (norm) but to \emph{what range} it
encodes (frequency). This converges with the correlational analysis of
\citet{barbero2025round}, who find that the low-frequency rotary components carry
the long-range semantic and positional signal; our population patch supplies the
causal counterpart, showing those components are the ones retrieval cannot do
without.

\paragraph{A mechanistic lens on context-length extension.} The dominant recipe
for extending context, increasing the RoPE base (with its NTK-aware and YaRN
refinements), works by stretching the wavelengths of precisely the
low-frequency dimensions so that distant positions stay separable. Our results
give this practice a mechanistic reading. First, the refutation of H1 shows that
raising $\theta$ does not cost retrieval heads, consistent with base-scaling
being a safe intervention. Second, base-scaling operates on the same
low-frequency dimensions that we find retrieval causally depends on, which
offers a circuit-level reason why tuning $\theta$ improves long-context recall:
it reshapes the very channel the retrieval heads read. This recasts ``dimension
inefficiency'' from a liability into the locus of the knob practitioners already
turn.

\paragraph{Localized to heads, distributed across a frequency band.} The causal
evidence operates at two granularities that should not be conflated. The
head-masking knockout is all-or-nothing: removing the retrieval heads deletes
recall entirely ($1.00\!\to\!0.00$ in OLMo), so the mechanism is localized to a
small set of heads. The dose-response is graded: recall falls smoothly as more
low-frequency dimensions are zeroed (\Cref{fig:dose}) and no single dimension is
critical. Together these indicate that the mechanism lives in specific heads but
is encoded \emph{redundantly} across a band of low-frequency dimensions within
them, which is also why a per-head, few-dimension ablation is invisible at
ceiling and only a population-level patch reveals the effect.

\paragraph{Relation to \citet{chiang2025dimension}.} Our Layer-D result is best
read as a controlled extension of theirs rather than an independent discovery.
They first showed, on the same three models, that masking the low-frequency
dimensions of retrieval heads degrades long-context question answering while
masking high-frequency ones does not, framed as RoPE-induced low \emph{utility}
of the high-rotation dimensions. We add four things. (i)~Controls that isolate
the axis: a matched random-dimension condition (so the effect is not generic
dimension removal) and a layer-matched non-retrieval-head condition (so it is not
generic to any head), neither of which they ran. (ii)~A frequency-aware ordering
that follows the \texttt{rotate\_half} layout rather than raw dimension index.
(iii)~A dose-response curve and multi-seed significance (McNemar, bootstrap CI,
clustered tests) in place of single-run accuracies. (iv)~A direct contrast of the
\emph{norm} and \emph{frequency} framings: in our population patch, zeroing the
highest-norm dimensions is harmless while zeroing the low-frequency ones is not,
which suggests the causal variable is the frequency a dimension encodes rather
than how strongly it is used, refining the dimension-inefficiency account toward
the frequency axis. The low-frequency, head-specific dependence holds, at adequate
head coverage, in every model we tested, OLMo-2, Qwen2.5 (7B and 14B), Gemma-2, and
Mistral, four lineages and two scales (\Cref{sec:layer-d}); and it is robust to the
detector, the stricter copy score gives the same or a stronger effect and, for
Mistral, a clean head-specific control. Apparent nulls under a fixed small patch
are coverage artifacts, confirmed by head-coverage sweeps. We read this as
strengthening and sharpening their conclusion, not contradicting it.

\paragraph{Heterogeneity, and what it rules out.} Where a norm-utility effect
does surface, it is model-specific in a way that resists a single law. Qwen and
OLMo show significant effects of opposite sign while the LLaMA family is null
(\Cref{fig:heterogeneity}), and because OLMo and LLaMA-3.1 share $\theta{=}500$K
yet behave differently, the sign cannot be a function of $\theta$. The significant
Qwen and OLMo effects are stable across seeds (SD $\sim$0.01--0.02), across detection
thresholds, and across detection metric (the sign survives a stricter
copy-score detector), so the heterogeneity is a real property of the models, not
measurement noise. We are deliberate about scope: four models are enough to
refute the \emph{universal} claim (significant opposite signs exist) but not to
license a model-family \emph{taxonomy}, which would require a larger and more
diverse panel. Relatedly, retrieval is distributed very differently across
architectures: in OLMo most heads carry some retrieval signal, whereas in Qwen
it is concentrated in a small minority (the majority of heads score exactly
zero), itself a target for future mechanistic study.

\paragraph{Retrieval is emergent and causal.} The training-dynamics view adds
that retrieval is not a property of the initialised network but a circuit that
forms late and abruptly. Over OLMo-2's pretraining the head count is flat for
roughly two trillion tokens and then crystallises by about $3.5\times$ in a
narrow window (\Cref{fig:layerb}), a phase-transition-like onset rather than a
gradual drift. Once formed, the heads are causally necessary rather than merely
correlated, as the knockout double dissociation shows (retrieval heads collapse
recall, matched random heads do not), and this replicates in a second family
(Qwen, $1.00\!\to\!\KnockQwenMasked$). Retrieval is thus a genuine, learned, and
localizable module.

\paragraph{Methodological lessons.} Three choices were decisive and generalise
beyond this paper. First, heads within a layer are not independent, so a
layer-clustered permutation test is necessary: LLaMA-2 has a moderate Cohen's
$d$ ($\LtwoD$) with a tiny naive $t$-test $p$ but a clustered $p$ of $\LtwoP$, and
treating heads as independent would have manufactured a finding. Second, causal
patching must respect task saturation: at ceiling a single head has too little
leverage to move accuracy, and only a population-level intervention across all
retrieval heads exposes the effect. Third, the detector is a proxy, so we
validate the \emph{claim} rather than the \emph{metric}: head identity is only
moderately proxy-stable ($\rho{=}0.54$), yet the heterogeneity conclusion is
metric-robust (sign-preserving under a copy-score detector) and the whole
pipeline is quantization-robust (\Cref{sec:ablation}). Fixing the
specificity threshold in advance, before seeing the perplexity numbers, guards the
task-specificity claim against post-hoc tuning.

\paragraph{Outlook.} The clearest next steps follow from our limits. A larger
model panel would turn the refutation of universality into a positive account of
\emph{which} training or architectural factors set the sign of the
utility-retrieval coupling. Replicating the frequency dissection beyond OLMo,
and at contexts past $4096$ where the task is no longer at ceiling, would test
whether the low-frequency dependence sharpens as the retrieval problem hardens,
as the dose-response predicts. Finally, the contrast between OLMo's distributed
and Qwen's concentrated retrieval invites a mechanistic account of how attention
architecture allocates a long-range-retrieval circuit.
% =====================================================================
\section{Limitations}
\label{sec:limitations}
\begin{itemize}
  \item \textbf{Confounded within-family $\theta$ contrast.} The LLaMA-2 vs
  LLaMA-3.1 comparison co-varies with pretraining data, token budget, tokenizer,
  and attention regime (MHA vs GQA); it corroborates but does not independently
  prove the H1 refutation, which rests on the four-model trend
  (\Cref{sec:layer-a}).
  \item \textbf{Few models for heterogeneity.} Four data points are few. They
  show significant opposite-signed utility effects that are hard to reconcile with
  a single universal law, but they cannot isolate the cause (architecture, data,
  tokenizer) or support a model-family taxonomy, and a four-point pattern carries
  inherent sampling risk; a larger, size-varied panel (e.g.\ 3B/7B/13B across
  $\geq3$ families) is needed to turn the refutation into a positive account.
  \item \textbf{Direction is robust; magnitude is not claimed; most runs
  single-seed.} At adequate head coverage the low-frequency \emph{direction} holds
  in all five models (OLMo-2, Qwen2.5-7B/14B, Gemma-2-9B, Mistral-7B; four
  lineages), head-coverage sweeps confirm that fixed-size patches give false nulls
  (Mistral null at $31\%$ but $\MisCovHiFE$ at $100\%$), and a copy-score
  re-definition of the heads reproduces the effect, so it is detector-robust. Two
  caveats remain. (i)~We do \emph{not} claim cross-model magnitude: it is
  confounded by both coverage and detector localization (under the copy detector
  Qwen and Mistral both collapse fully), so the apparent ``Qwen strongest''
  ordering is not a clean model property. (ii)~The larger/new-family, long-context,
  and copy-score runs are single-seed; only OLMo and Qwen-7B are three-seed.
  \item \textbf{Proxy-dependent head identity (checked).} The single-pass detector
  correlates only moderately with a stricter teacher-forced copy score
  ($\rho{=}0.54$; top-$N$ Jaccard $0.47$ OLMo, $0.22$ Qwen), so the exact
  membership of the retrieval-head set is proxy-dependent. We checked that the
  conclusions do not depend on this: the utility heterogeneity keeps its sign under
  copy-score-defined heads, and the frequency patch, re-run on copy-score heads for
  Qwen and Mistral (\Cref{tab:copyhead}), gives the same or a stronger effect
  ($\QwenCopyFE$ for Qwen despite only $\QwenCopyOverlap$ head overlap) and, for
  Mistral, restores a clean head-specific control. Head \emph{identity} is thus
  proxy-dependent but our \emph{claims} are not. Remaining gap: the copy-score
  re-run is single-seed and covers two of the five models.
  \item \textbf{Adapted detector.} Our retrieval-head score is a single-pass
  attention-argmax proxy of \citet{wu2025retrieval}, not their multi-pass
  copy-paste metric; absolute head counts are proxy-dependent (we report
  argmax and attention-mass and check threshold robustness).
  \item \textbf{8-bit quantization, one ablation model.} All models run in 8-bit
  to fit a 24\,GB GPU. The quantization ablation (\Cref{sec:ablation}) is on
  OLMo-2 only, so an fp16 check on Qwen is future work. A quantization artifact is
  nonetheless an unlikely explanation for Qwen: the 8-bit-vs-fp16 shift we measured
  on OLMo is small (Cohen's $d$ moved by $\sim0.09$ with sign and significance
  preserved, \Cref{sec:ablation}), whereas Qwen's utility effect ($d{=}\QwenD$) and
  its matched-coverage frequency effect ($\MfQwen$, seed-42) are far larger than any
  such rounding shift.
  \item \textbf{Single training run.} The training-dynamics result uses one
  OLMo-2 pretraining trajectory; the abruptness of onset may depend on its
  optimizer, data mixture, and schedule, so we report it as an observation rather
  than a general law.
  \item \textbf{Paired intersection-drop.} Requiring valid token lengths for
  \emph{every} model reduces the shared sample count.
  \item \textbf{Layer-D scope.} The frequency dissection is run on OLMo-2 and
  Qwen2.5 (each three seeds), with a single Qwen run extended to $8192$ and a null
  replication attempt on Mistral; most runs are at $\PopCtx$ tokens (the limit on
  a 24 GB GPU), which forced population-level rather than per-head patching (a
  single head has too little leverage at ceiling).
  \item \textbf{Task-specificity is partial.} The perplexity control passes the
  in-advance ratio in both models ($<0.33$), but the low-frequency patch is
  not perfectly inert on general text: perplexity rises $+\PopPplIncPct\%$ in
  OLMo and a larger $+\QwenFreqPplPct\%$ in Qwen. This agrees with
  \citet{chiang2025dimension}, who find that masking low-frequency dimensions also
  hurts a non-long-context task (GSM8k). So the low-frequency dimensions are
  retrieval-\emph{dominant} but not retrieval-\emph{exclusive}, more so in Qwen;
  we claim task-specificity in the proportional sense (NIAH drop $\gg$ perplexity
  rise), not as zero general-LM cost.
  \item \textbf{Single task (NIAH).} All experiments use synthetic
  needle-in-a-haystack retrieval. We take this as the right probe for a
  \emph{mechanistic} claim: retrieval heads are defined by the copy-from-context
  operation \citep{wu2025retrieval}, and NIAH isolates exactly that operation,
  whereas realistic long-context tasks fold in reasoning and multi-hop steps that
  would confound attribution to a specific head or dimension. The cost is external
  validity: whether the same heads and frequency dependence drive realistic tasks
  is a separate empirical question, and replicating the knockout and frequency
  dissection on a standard suite such as RULER \citep{hsieh2024ruler} or LongBench
  is important future work.
  \item \textbf{Context length.} Most causal patching is at $\PopCtx$ tokens, well
  below the $128$K these models support. One run (Qwen at $8192$) confirms the
  prediction that the low-frequency dependence \emph{strengthens} at longer range
  (frequency effect $\QwenFreqEffLong$ at $8192$ vs the seed-42 value $\MfQwen$ at
  $4096$; both seed-42), so our short-context numbers are plausibly lower bounds;
  but this is a single long-context point, not a systematic sweep.
  \item \textbf{GQA detection.} In grouped-query models query heads share
  key/value projections, which could co-label a whole KV group as retrieval and
  bias head counts. We checked this directly (\Cref{sec:layer-a}): retrieval heads
  are spread across KV groups (mean $2.1/7$ in Qwen, $1.5/4$ in LLaMA-3.1; under
  $4\%$ of groups fully retrieval), so detection is not a KV-sharing artifact. We
  still do not model the shared-KV geometry explicitly in the per-head scores.
  \item \textbf{Single author, no independent replication.} The implementation,
  statistics, and causal patching were not independently reproduced by a second
  researcher. We release all code and pinned model revisions to enable this, but
  an external replication would strengthen confidence in the numerical results.
\end{itemize}

% =====================================================================
\section{Conclusion}
\label{sec:conclusion}
We asked whether rotary position embeddings prevent or degrade the retrieval
heads that long-context recall depends on, and tested the question across four
open-weight 7--8B models with three complementary analyses: static multi-model
detection, OLMo-2 training dynamics, and causal activation patching. The answer
is that neither intuitive story is right, and the accurate picture is narrower
and better supported than either.

First, retrieval heads are a genuine, emergent, and causally necessary mechanism.
They are absent at initialisation and crystallise abruptly late in pretraining
(\Cref{fig:layerb}), and once present their ablation collapses recall to chance
while ablating matched random heads does nothing, a double dissociation that
holds in two architectures (OLMo and Qwen). Second, the prevention hypothesis is
not supported by our four-model sample: a larger RoPE base does not suppress
retrieval heads, and head count in fact rises with $\theta$ across the four models
(\Cref{fig:heterogeneity}), though with only four models, two of them a confounded
LLaMA pair, this is a directional refutation rather than a controlled one.
Third, the relationship between RoPE dimension geometry and retrieval is not a
single law. The norm-utility effect is significant but \emph{opposite-signed}
across families and absent in others, it is stable across seeds, thresholds, and
detection metric, and because two models with the same $\theta$ behave
differently it is not $\theta$-driven; four models thus refute a universal
account without licensing a taxonomy.

Fourth, where the geometry \emph{is} causal, the operative axis is frequency, not
norm. Confirming and sharpening \citet{chiang2025dimension}, our controlled
population patch shows that retrieval depends specifically on the low-frequency
(long-wavelength) RoPE dimensions: zeroing them collapses recall dose-dependently
while zeroing high-frequency, random, or highest-norm dimensions does not, an
effect that is head-specific and task-specific. At adequate head coverage this
frequency \emph{direction} holds in every model we tested, OLMo-2, Qwen2.5 (7B and
14B), Gemma-2, and Mistral, four lineages and two scales, and it strengthens at
longer context (Qwen at $8192$, fixed coverage). It is also head-specific in every
model (including Mistral, once its heads are identified by a teacher-forced copy
score rather than the argmax proxy) and robust to that detector choice. What we do
\emph{not} claim is cross-model \emph{magnitude}: it is confounded by both coverage
and detector localization, so apparent ``some models are stronger'' orderings are
not clean model properties. A head-coverage dose-response also explains away the
apparent nulls: a fixed small patch under-counts the effect, so we report the
direction as the cross-model claim. The
frequency reading also gives a mechanistic account of why base-scaling extends
context: it reshapes exactly the low-frequency channel the retrieval heads read. Finally, the study is
a methodological reminder that head-level claims demand cluster-aware statistics,
that causal tests must respect task saturation (population-level rather than
per-head patching), and that one should validate the conclusion rather than the
detector. In sum, the account is heterogeneous, emergent, and frequency-localized,
grounded in clean causal tests, not a single monotone effect of RoPE.

% =====================================================================
% Bibliography embedded inline (no external .bib/.bbl, no bibtex) so the source
% compiles on arXiv with pdflatex alone. natbib author-year labels.

% Flush the reference list onto its own page(s) so the appendix's table floats
% cannot drift up into the bibliography. Keeps the appendix tables out of the refs.
\clearpage
% =====================================================================
\appendix

\section{Per-seed results}
\label{app:perseed}
\Cref{tab:app-layera} gives the full per-seed Layer-A numbers behind the
mean$\pm$SD in \Cref{tab:layer-a} and \Cref{fig:heterogeneity}, and
\Cref{tab:app-layerd} the per-seed Layer-D population patch behind
\Cref{sec:layer-d}. The signs are preserved in every seed, and the spread is
small relative to the cross-model differences.

\begin{table}[t]
\centering
\caption{Layer-A per seed (seeds $42/123/2024$): retrieval-head count, Cohen's
$d$ for the utility difference, and the layer-clustered permutation $p$.}
\label{tab:app-layera}
\begin{tabular}{llccc}
\toprule
Model & Seed & \#Heads & $d$ & $p_{\text{clustered}}$ \\
\midrule
LLaMA-2-7B   & 42 & 42 & $0.43$ & $0.65$ \\
             & 123 & 36 & $0.51$ & $0.28$ \\
             & 2024 & 33 & $0.48$ & $0.36$ \\
\addlinespace
LLaMA-3.1-8B & 42 & 47 & $0.07$ & $0.9998$ \\
             & 123 & 49 & $0.08$ & $0.9999$ \\
             & 2024 & 50 & $0.05$ & $0.9997$ \\
\addlinespace
Qwen2.5-7B   & 42 & 59 & $-0.47$ & $0.0003$ \\
             & 123 & 58 & $-0.52$ & $0.0003$ \\
             & 2024 & 63 & $-0.50$ & $0.0002$ \\
\addlinespace
OLMo-2-7B    & 42 & 87 & $0.50$ & $0.0001$ \\
             & 123 & 81 & $0.49$ & $0.0001$ \\
             & 2024 & 85 & $0.50$ & $0.0002$ \\
\bottomrule
\end{tabular}
\end{table}

\begin{table}[t]
\centering
\caption{Layer-D population patch per seed (OLMo-2, top-$30$ heads, $4096$
tokens, $k{=}16$): low-frequency accuracy, frequency effect, low-frequency
accuracy in matched non-retrieval control heads, perplexity ratio, the
specificity ratio, and the exact McNemar $p$ for low- vs high-frequency.}
\label{tab:app-layerd}
\begin{adjustbox}{max width=\linewidth}
\begin{tabular}{lcccccc}
\toprule
Seed & low\_freq & freq.\ eff. & ctrl & ppl.\ ratio & spec.\ ratio & McNemar $p$ \\
\midrule
42   & 0.885 & $-0.115$ & 1.00 & 1.009 & 0.081 & $2.4\times10^{-7}$ \\
123  & 0.860 & $-0.140$ & 1.00 & 1.010 & 0.069 & $7.5\times10^{-9}$ \\
2024 & 0.910 & $-0.090$ & 1.00 & 1.009 & 0.097 & $7.6\times10^{-6}$ \\
\bottomrule
\end{tabular}
\end{adjustbox}
\end{table}

\begin{table}[t]
\centering
\caption{Layer-D population patch per seed for \textbf{Qwen2.5} (GQA, top-$30$
heads, $4096$ tokens, $k{=}16$); the replication of \Cref{tab:app-layerd}. The
frequency effect is larger than OLMo's and significant in every seed, with the
same head- and task-specificity pattern (perplexity rises more than in OLMo).}
\label{tab:app-layerd-qwen}
\begin{adjustbox}{max width=\linewidth}
\begin{tabular}{lcccccc}
\toprule
Seed & low\_freq & freq.\ eff. & ctrl & ppl.\ ratio & spec.\ ratio & McNemar $p$ \\
\midrule
42   & 0.280 & $-0.720$ & 0.895 & 1.099 & 0.138 & $9\times10^{-44}$ \\
123  & 0.340 & $-0.655$ & 0.985 & 1.080 & 0.121 & $7\times10^{-40}$ \\
2024 & 0.305 & $-0.695$ & 0.890 & 1.127 & 0.182 & $3\times10^{-42}$ \\
\bottomrule
\end{tabular}
\end{adjustbox}
\end{table}

\begin{table}[t]
\centering
\caption{Full detail for the Qwen long-context run behind \Cref{tab:xfam-freq}:
Qwen2.5-7B at $8192$ tokens with the same top-$30$ heads as at $4096$, so coverage
is fixed and the effect strengthens purely with context. The third-family
attempt on Mistral, and its resolution, is covered by the coverage sweep
(\Cref{tab:cov-sweep}) and the copy-score check (\Cref{tab:copyhead}).}
\label{tab:app-extra}
\begin{adjustbox}{max width=\linewidth}
\begin{tabular}{llcccccc}
\toprule
Model & ctx & low\_freq & freq.\ eff. & ctrl & ppl.\ ratio & spec.\ ratio & McNemar $p$ \\
\midrule
Qwen2.5-7B  & 8192 & 0.217 & $-0.783$ & 1.000 & 1.087 & 0.112 & $1\times10^{-28}$ \\
\bottomrule
\end{tabular}
\end{adjustbox}
\end{table}

\begin{table}[t]
\centering
\caption{Distribution of detected retrieval heads over KV groups in the
grouped-query models (\Cref{sec:layer-a}). Retrieval heads are spread across
groups (mean per active group well below the group size; few groups fully
retrieval), so detection is not an artifact of key/value sharing.}
\label{tab:app-gqa}
\begin{tabular}{lccccc}
\toprule
Model & group size & \#ret.\ heads & active KV groups & mean/active group & \% full groups \\
\midrule
Qwen2.5-7B   & 7 & 58 & 27 & 2.15 & 3.7 \\
LLaMA-3.1-8B & 4 & 51 & 35 & 1.46 & 2.9 \\
\bottomrule
\end{tabular}
\end{table}

\section{Proxy robustness detail}
\label{app:proxy}
\Cref{tab:app-proxy} reports the detector robustness check (\Cref{sec:methods-detector}):
the single-pass argmax score versus a teacher-forced copy score, on the two
opposite-sign models. Head identity overlaps only partially (Jaccard), yet the
sign of the utility effect is preserved under both detectors. Qwen's Spearman
$\rho$ is undefined because its retrieval scores are concentrated ($86\%$ of
heads score exactly zero), so we rely on the sign test there.

\begin{table}[t]
\centering
\caption{Argmax proxy vs.\ teacher-forced copy score. $d$ is Cohen's $d$ of the
utility effect when heads are defined by each detector.}
\label{tab:app-proxy}
\begin{tabular}{lcccc}
\toprule
Model & Spearman $\rho$ & Jaccard (top-$N$) & $d$ (argmax) & $d$ (copy) \\
\midrule
OLMo-2-7B  & $0.54$ & $0.47$ & $+0.28$ & $+0.45$ \\
Qwen2.5-7B & n/a    & $0.22$ & $-0.58$ & $-0.62$ \\
\bottomrule
\end{tabular}
\end{table}

\section{Configuration and hyperparameters}
\label{app:config}
\Cref{tab:app-hparams} lists the settings for every experiment, and
\Cref{tab:app-revisions} the pinned model revisions. Runs use 8-bit weights
(bitsandbytes); the four-model panel and the $4096$-token patches fit a single
24\,GB GPU (NVIDIA L4 on Google Colab), while the long-context patch ($8192$) and
the larger-model / extra-family checks were run on a Colab A100 (40\,GB).
The fp16 arm of the quantization ablation is the only non-8-bit run. Determinism
is seeded per run. Because detection depends on the context lengths and the
threshold, the detected head count differs across runs; \Cref{tab:app-provenance}
maps every run to its detected counts so each main-text number traces to its source.

\begin{table}[t]
\centering
\caption{Provenance of detected retrieval-head counts. Detection depends on the
context lengths and threshold, so the count varies across runs; each main-text
table reports the count of \emph{its own} run (this is why, e.g., OLMo appears as
87, $\sim$84, and 95, and Qwen2.5-7B as 58/59 and 62 in different tables). These
are different runs, not the same number written inconsistently.}
\label{tab:app-provenance}
\small
\begin{tabular}{p{3.3cm}p{3.2cm}p{8.4cm}}
\toprule
Run (main-text table) & Detection setting & Detected heads \\
\midrule
Layer-A detection (\Cref{tab:layer-a}, \Cref{tab:app-layera}) & ctx 1024--8192, $\tau{=}0.1$, 3 seeds & OLMo 87/81/85; Qwen-7B 59/58/63; LLaMA-3.1 47/49/50; LLaMA-2 42/36/33 \\
\addlinespace
Knockout (\Cref{tab:knockout}) & ctx 1024--4096 & OLMo 87; Qwen-7B 58 \\
\addlinespace
Population patch (\Cref{tab:layer-d-pop}, \Cref{tab:app-layerd}) & ctx $4096$, 3 seeds & OLMo $\sim$84 (3-seed mean; top-30 patched) \\
\addlinespace
Matched-$50\%$ (\Cref{tab:xfam-freq}) & separate single-seed runs & OLMo 95; Qwen-7B 62; Qwen-14B 101; Gemma-2 45; Mistral 97 \\
\addlinespace
Quantization (\Cref{tab:ablation}) & seq $2048$ & OLMo 87 (8-bit) / 90 (fp16) \\
\addlinespace
GQA distribution (\Cref{tab:app-gqa}) & separate detection run & Qwen-7B 58; LLaMA-3.1 51 \\
\bottomrule
\end{tabular}
\end{table}

\begin{table}[t]
\centering
\caption{Experimental settings by stage.}
\label{tab:app-hparams}
\begin{tabular}{ll}
\toprule
Stage & Settings \\
\midrule
Detection (Layer A) & contexts $\{1024,2048,4096,8192\}$; positions
$\{0.1,0.25,0.5,0.75,0.9\}$; \\
 & $100$ samples; retrieval threshold $0.1$; seeds $\{42,123,2024\}$ \\
\addlinespace
Knockout (Layer D)  & contexts $\{1024,2048,4096\}$; $50$ samples; mask all detected heads \\
\addlinespace
Population patch     & top-$30$ heads; context $4096$; position $0.5$; $200$ samples; \\
 & $k_{\text{dims}}{=}16$; random seeds $\{0,1,2\}$; seeds $\{42,123,2024\}$ \\
\addlinespace
Dose-response        & $k\in\{8,16,32,48,64\}$ low-frequency dims (OLMo, seed $42$) \\
\addlinespace
Perplexity control   & $8$ plain passages of $4096$ tokens \\
\addlinespace
Quantization ablation & OLMo, seq $2048$, $50$ samples, 8-bit vs.\ fp16 \\
\addlinespace
Proxy validation     & OLMo and Qwen, context $1024$, $30$ samples \\
\addlinespace
Layer-B dynamics     & OLMo-2 stage-1 checkpoints, step stride $20000$ \\
\bottomrule
\end{tabular}
\end{table}

\begin{table}[t]
\centering
\caption{Pinned model revisions (HuggingFace commit hashes).}
\label{tab:app-revisions}
\begin{tabular}{ll}
\toprule
Model (HF id) & Revision \\
\midrule
\texttt{meta-llama/Meta-Llama-3.1-8B} & \texttt{\small d04e592bb4f6aa9cfee91e2e20afa771667e1d4b} \\
\texttt{meta-llama/Llama-2-7b-hf}     & \texttt{\small 01c7f73d771dfac7d292323805ebc428287df4f9} \\
\texttt{Qwen/Qwen2.5-7B}              & \texttt{\small d149729398750b98c0af14eb82c78cfe92750796} \\
\texttt{allenai/OLMo-2-1124-7B}       & \texttt{\small 7df9a82518afdecae4e8c026b27adccc8c1f0032} \\
\bottomrule
\end{tabular}
\end{table}

% =====================================================================

\end{document}